\documentclass{article}
\usepackage{iclr2026_conference,times}
\usepackage[T1]{fontenc}
\usepackage[utf8]{inputenc}
\usepackage{amsmath,amssymb}
\usepackage{microtype}
\usepackage{graphicx,booktabs,enumitem}
\usepackage[colorlinks=true,citecolor=blue,linkcolor=blue,urlcolor=blue]{hyperref}
\usepackage{url}
\makeatletter
\renewcommand{\paragraph}{\@startsection{paragraph}{4}{\z@}{1ex plus .5ex minus .2ex}{-1em}{\normalsize\bf}}
\makeatother

\newcommand{\upshift}{\textsc{Upshift}}
\title{\upshift{}: User Preference Steering from\\Human--AI Interaction Feedback Traces}
\author{%
  \href{https://openreview.net/profile?id=~Jingping_Wang1}{Jingping Wang}$^{1,2}$\quad
  \href{https://openreview.net/profile?id=~Lizhanhao1}{Lizhanhao}$^{2}$\quad
  \href{https://openreview.net/profile?id=~Shao_Zhang1}{Shao Zhang}$^{2}$\quad
  \href{https://openreview.net/profile?id=~Shuyue_Hu1}{Shuyue Hu}$^{2}$\\[0.4em]
  \normalfont $^{1}$Shanghai Jiaotong University (\url{sjtu.edu.cn})\\
  \normalfont $^{2}$Shanghai Artificial Intelligence Laboratory (\url{shlab.org.cn})
}
\iclrfinalcopy
\hypersetup{
  pdftitle={UPSHIFT: User Preference Steering from Human-AI Interaction Feedback Traces},
  pdfauthor={Jingping Wang, Lizhanhao, Shao Zhang, Shuyue Hu}
}

\begin{document}
\maketitle
% Preprint: no anonymous-review or conference-acceptance statement.
\fancyhead{}
\begin{abstract}
Personalizing large language model assistants requires adapting their responses to individual users, including the selection of content, reasoning approach, level of detail, and form of presentation. Such preferences are often latent, becoming visible only when users respond to an initial answer and request revisions. We investigate whether these feedback trajectories can be converted into a reusable personalization signal without explicit preference profiles or labels, prompt-time access to interaction histories, or model parameter updates. We introduce \upshift{} (User Preference Steering from Human--AI Interaction Feedback Traces), a training-free method that compares the hidden representations of initial and feedback-conditioned responses within each session, averages the resulting trajectory displacements across a user's sessions, and injects the aggregated preference direction when answering a new query. The method is motivated by the implicit-preference hypothesis: although an individual conversational transition may reflect topic progression or task-specific correction, changes that recur across sessions retain a component that captures how the user wants responses to be adjusted. We validate the preference-representation capability of \upshift{} on PrefEval. In evaluations with 40 simulated users, every tested configuration improves preference alignment over the unsteered baseline on held-out topics for both Qwen3-8B and Qwen3-32B; the best configuration increases macro-averaged Elo from 1364.0 to 1605.2 and from 1341.9 to 1591.8, respectively. Matched-user and positive-versus-negative controls further demonstrate that these gains arise from the user-specific preference direction and its learned orientation. These results show that interaction histories can provide compact, reusable supervision for lightweight personalization.

\end{abstract}
\section{Introduction}
\label{sec:introduction}

Large language model assistants increasingly serve users who differ not only in what they ask, but also in what they consider a useful response. Even when factuality and safety requirements are fixed, users may prefer different facts to be emphasized, different amounts of reasoning to be shown, different treatments of uncertainty, or different forms of presentation. Personalization must therefore adapt how the assistant responds rather than merely imitate how the user writes. The challenge is that these preferences are often latent, context-dependent, and difficult to specify in advance. Some become apparent only after the user sees a response and decides what should change: a terse prompt does not imply a preference for a terse answer, just as an informal message does not imply a preference for an informal report.

Existing language-model personalization methods typically represent users through explicit profiles, preference labels, retrievable histories, or collections of user-authored text. These representations are useful when preferences can be named, summarized, or directly reused in context, but much of the evidence in interaction is less explicit. A user may never state a general rule such as ``I prefer derivations before intuition,'' yet repeatedly ask for missing steps, redirect high-level explanations toward formal arguments, or remove presentation choices that obscure the reasoning. Such interventions do not necessarily describe a preference, but they indicate how the current response should move to better satisfy the user. Although a full dialogue history preserves this evidence, converting it into an explicit, prompt-ready preference description can be difficult, and a natural-language summary may express only part of the response behavior the user repeatedly requests.

We investigate whether a user-specific steering vector can serve as a compact preference controller inferred implicitly from historical human--AI interactions. Unlike profile-based approaches, this route does not first translate the interaction history into an explicit, prompt-level preference description. Instead, it estimates the controller directly from changes in the model's hidden representations over the course of a dialogue. Feedback-bearing turns---such as ``make it shorter,'' ``show the derivation,'' or ``do not use a table''---provide observations of how the user wants the assistant's response to change. By aggregating the resulting representation displacements across conversations, we aim to recover a recurring preference direction that can steer responses to new queries without placing the original history in context.

This approach rests on a statistical assumption about conversational feedback. A single feedback-conditioned transition may reflect task progression, a topic shift, a factual correction, or even an unsuccessful revision rather than a stable user preference. We hypothesize, however, that across multiple feedback-bearing interactions from the same user, these session-specific changes vary while a recurring component reflecting how the user wants responses to change persists under aggregation. We refer to this as the implicit-preference hypothesis.

\begin{figure}[t]
  \centering
  \includegraphics[width=\linewidth]{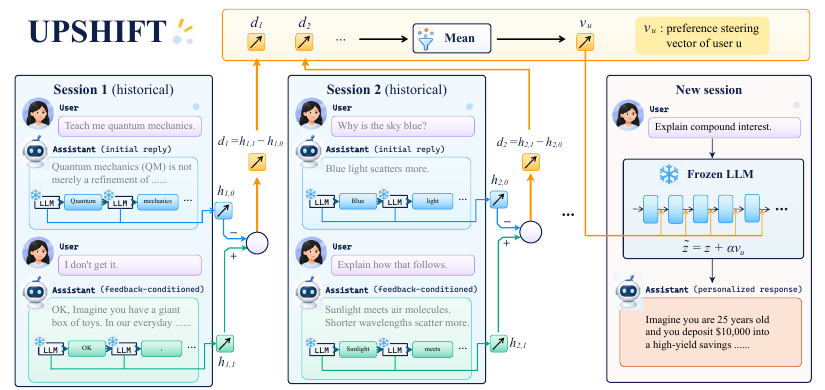}
  \caption{\textbf{Overview of \upshift{}.} Within each historical session $s$, we extract hidden representations from the initial reply and the feedback-conditioned reply and compute their displacement, $d_s=h_{s,1}-h_{s,0}$. We average these displacements across the user's sessions to obtain a reusable preference direction $v_u$. For a new query, we add the scaled direction $\alpha v_u$ to the residual stream of the frozen language model, steering generation toward a response aligned with the user's revealed preferences without updating model parameters.}
  \label{fig:overview}
\end{figure}

We instantiate this idea as \upshift{}, a training-free personalization method built on activation steering \citep{turner2023,zou2023,panickssery2023}. Given a frozen language model, \upshift{} records hidden representations for the initial and selected feedback-conditioned replies, treating their difference as a trajectory displacement. It aggregates one such displacement per session into a single direction and adds that direction to the model's residual stream when answering a new query. The method requires neither gradient optimization nor a natural-language user model. Recent work has also used activation-space representations for personalization, for example by deriving style vectors from user-authored histories or synthetic style-neutral counterfactuals, or by exposing predefined preference controls \citep{zhang2025,bo2026,zhaox2025,chen2026}. \upshift{} differs in both objective and signal: it targets assistant-side preference alignment rather than authorial imitation, and extracts the temporal displacement within ordinary human--AI dialogue rather than relying on a named preference, explicit rating, synthetic negative example, textual profile, or collection of target writing samples. Figure~\ref{fig:overview} gives an overview.

The central scientific question is not merely whether adding an activation vector changes model outputs, which a sufficiently large intervention almost always will. It is whether the resulting change (i) is preferred by the same user, (ii) transfers beyond the topics used to estimate the direction, and (iii) preserves general response quality. We design our evaluation around these criteria and include controls for generic turn-depth effects, topic leakage, response length, and judge position bias.

Our contributions are:
\begin{itemize}
\item We introduce \upshift{}, a training-free temporal-contrast estimator that implicitly derives a reusable user-specific steering vector from natural human--AI interaction histories.
\item We formulate an implicit-preference hypothesis that motivates the estimator: user interventions are noisy but directionally informative observations of how the user wants an assistant to respond.
\item We combine cross-session aggregation with cross-topic evaluation to distinguish persistent user preference from session-specific and local semantic content.
\item We provide a falsifiable evaluation protocol, including negative-direction interventions, random and other-user controls, preference-strength scaling, and capability-preservation tests.
\end{itemize}

\section{Related Work}
\label{sec:related}

\paragraph{Personalized language models.} Steering language models to produce personalized responses that better match individual user preferences is an active area of research. Most approaches to preference adaptation first construct an explicit representation of the user and then condition the model on that representation. The representation may consist of prior documents, retrieved interaction histories, or a natural-language user profile \citep{salemi2024,tang2024,jiang2025,guo2026}. Such representations work well when preferences can be summarized explicitly or when a relevant part of the user's history can be retrieved for a new request. However, interaction histories often do not contain preference descriptions that can be extracted directly, and many of these methods require user-specific information to be included in the input context.

User edits provide a more direct source of evidence. PRELUDE and CIPHER infer natural-language preference descriptions from user-edited model outputs and include these descriptions in subsequent prompts \citep{gao2024}.

\upshift{} starts from the same broad idea that user interventions can reveal preferences. However, it neither constructs a textual user profile nor retrieves historical examples at inference time. Instead, it compresses the preference-related changes that recur across conversations into a fixed-dimensional activation vector.

\paragraph{Learning from conversational feedback.} Conversations between users and language models contain corrections, revision requests, and other forms of feedback. These interactions can provide supervision without requiring a separate annotation interface. Prior work mines such feedback to construct training data or optimize model parameters \citep{donyehiya2024,chen2025,liu2025,shi2026}.

These studies show that conversational feedback is abundant and useful, but also ambiguous and noisy. Our approach keeps feedback associated with the individual user instead of pooling it into a global training signal. It adapts the model at inference time without updating model parameters. Because implicit feedback is noisy, we aggregate evidence across interactions and use controlled comparisons to test whether the resulting steering direction captures a stable user preference.

\paragraph{Activation steering.} Activation Addition steers a pretrained language model by adding a chosen vector to its internal activations, thereby shifting generation toward a target behavior \citep{turner2023,zou2023}. Contrastive Activation Addition (CAA) estimates such a steering direction by averaging representation differences between labeled pairs of behavioral examples \citep{panickssery2023}.

The closest technical precedents to our work apply activation steering to personalization. StyleVector contrasts a user's historical writing with model-generated, style-neutral counterparts \citep{zhang2025}; SteerX identifies preference-related tokens when constructing a steering signal \citep{zhaox2025}; and GLASS combines global style features with local semantic calibration \citep{chen2026}.

Related work on steering interfaces further suggests that users value direct control over steering strength \citep{bo2026}.

Taken together, these studies show that activation-space personalization is feasible. However, reliable transfer across tasks and composition across multiple traits remain open challenges \citep{subbiah2026}.

\upshift{} builds on this line of research but changes both the optimization target and the source of supervision. Its goal is not to reproduce the user's writing style, but to adjust how the assistant responds so that the response better matches the user's preferences. The supervision comes from the temporal progression of natural dialogue: we contrast representations after later user interventions with the representation after the initial request to estimate the direction in which the user repeatedly steers the interaction. This temporal contrast provides a preference-oriented direction without requiring predefined preference dimensions, a collection of target writing samples, style-neutral counterexamples, or manually labeled positive--negative pairs.

\section{Problem Setup}
\label{sec:setup}

Let $M$ be a frozen decoder-only language model with residual width $d$ and layers $1,\ldots,L$. For a user $u$, the available history is a set of sessions
\begin{equation}
  \mathcal{D}_u=\{\tau_s\}_{s=1}^{S_u},\qquad
  \tau_s=(x_{s,0},y_{s,0},x_{s,1},y_{s,1},\ldots,x_{s,T_s}),
  \label{eq:history}
\end{equation}
where $x_{s,0}$ is an initial request, $y_{s,t}$ is an assistant response, and $x_{s,t}$ for $t>0$ is the user's next message. The goal is to use $\mathcal{D}_u$ to improve the response to a new query $x^*$ from the same user, without placing the full history in context and without modifying $M$'s parameters.

\paragraph{Implicit-preference hypothesis.} Let $R_u(y,x)$ denote the latent utility that user $u$ assigns to response $y$ for request $x$. This quantity is best understood as the user's subjective satisfaction with the response: it exists in the user's judgment, but neither the model nor an external personalization algorithm can observe its value or functional form directly. The user's behavior nevertheless provides indirect evidence about it. After seeing a response, the user chooses whether and how to intervene in order to move the model toward a response they expect to prefer. If $Y_{s,t-1}$ and $Y_{s,t}$ denote random responses drawn from the model immediately before and after a feedback-bearing intervention, respectively, the motivating behavioral premise is
\begin{equation}
  \mathbb{E}[R_u(Y_{s,t},x_{s,0})-R_u(Y_{s,t-1},x_{s,0})\mid x_{s,t}\text{ is feedback-bearing}]>0.
  \label{eq:utility}
\end{equation}

In words, a user's interventions are, on average, attempts to increase the satisfaction of the next response. Equation~\ref{eq:utility} does not require utility to increase monotonically in every observed conversation: the model may misunderstand the feedback, a follow-up may advance the task rather than revise it, and a user's goals may change. Our hypothesis is instead that, after aggregating many such noisy interventions, their induced representational displacements retain a stable component that points toward responses with higher utility for that user on future tasks. This is a positive-unlabeled assumption that we test directly rather than treat as ground truth.

\section{Method}
\label{sec:method}

\begin{figure}[t]
  \centering
  \includegraphics[width=\linewidth]{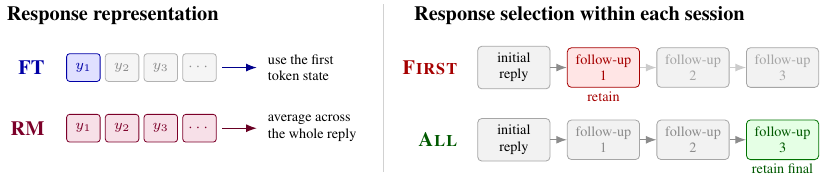}
  \caption{\textbf{The two independent choices in \upshift{}.} FT uses the first response-token state, whereas RM averages states across the reply. \textsc{First} uses the response after the first follow-up, whereas \textsc{All} uses the final response after all follow-ups.}
  \label{fig:choices}
\end{figure}

\upshift{} estimates a user-specific steering vector from hidden-state changes in historical conversations. Within each session, it contrasts a selected feedback-conditioned response with the initial response and aggregates the resulting displacement across the user's history. At each layer $\ell$ in a predefined candidate set $\mathcal{L}_{\mathrm{cand}}\subseteq\{1,\ldots,L\}$, the estimator makes two independent choices: whether to represent a response by its first-token state (FT) or its mean response state (RM), and whether to select the response after the first follow-up (\textsc{First}) or the final response after all follow-ups (\textsc{All}). This yields four configurations, illustrated in Figure~\ref{fig:choices}.

\subsection{Response representations}
\label{sec:representations}

Let $c_{s,t}$ be the dialogue prefix ending at user message $x_{s,t}$ in session $s$, rendered with the model's chat template. Let $y_{s,t}=(y_{s,t,1},\ldots,y_{s,t,n_{s,t}})$ be the assistant response associated with that prefix.

Here $t=0$ denotes the initial response, while $t\geq1$ denotes a response after one or more follow-ups; each prefix contains the preceding dialogue. Let $a^{\ell}_{s,t,k}\in\mathbb{R}^d$ be the hidden state at the $k$th assistant-response token at candidate layer $\ell$. Using only states within the assistant response, we define
\begin{equation}
  h^{\ell,p}_{s,t}=\begin{cases}
    a^{\ell}_{s,t,1}, & p=\mathrm{FT},\\[3pt]
    \displaystyle\frac{1}{n_{s,t}}\sum_{k=1}^{n_{s,t}}a^{\ell}_{s,t,k}, & p=\mathrm{RM}.
  \end{cases}
  \label{eq:pooling}
\end{equation}
The same pooling rule and activation site are used for all responses being contrasted.

\subsection{Baseline contrasts and aggregation}
\label{sec:aggregation}

For pooling rule $p\in\{\mathrm{FT},\mathrm{RM}\}$, the initial response $h^{\ell,p}_{s,0}$ is the session baseline. Each follow-up yields the baseline-relative displacement
\begin{equation}
  d^{\ell,p}_{s,t}=h^{\ell,p}_{s,t}-h^{\ell,p}_{s,0}.
  \label{eq:displacement}
\end{equation}
Thus, round $t$ is compared with round $0$, not with round $t-1$; later displacements capture the accumulated interaction. As discussed in Section~\ref{sec:setup}, they may also contain task- and session-specific changes, which aggregation aims to attenuate.

Let the user have $S_u$ sessions, with $T_s\geq1$ follow-up responses in session $s$. The two response-selection rules, followed by cross-session averaging, are
\begin{equation}
  \begin{aligned}
    v^{\ell,p}_{u,\mathrm{First}}&=\frac{1}{S_u}\sum_{s=1}^{S_u}d^{\ell,p}_{s,1},\\
    v^{\ell,p}_{u,\mathrm{All}}&=\frac{1}{S_u}\sum_{s=1}^{S_u}d^{\ell,p}_{s,T_s}.
  \end{aligned}
  \label{eq:aggregation}
\end{equation}
Both estimators weight sessions equally. \textsc{First} captures the change after the first intervention, whereas \textsc{All} captures the cumulative change at the end of the complete interaction.

\subsection{Steering a new response}
\label{sec:steering}

For a chosen configuration $(r,p)$, where $r\in\{\mathrm{First},\mathrm{All}\}$, the extraction procedure yields a layer-indexed collection of user directions $\{v^{\ell,p}_{u,r}:\ell\in\mathcal{L}_{\mathrm{cand}}\}$. We select an intervention layer $\ell^*\in\mathcal{L}_{\mathrm{cand}}$ and steering coefficient using validation data. Let $z_k^{\ell^*}$ be the activation at the selected intervention site during generation for a new query $x^*$. At each intervened decoding position, we apply
\begin{equation}
  \widetilde{z}_k^{\ell^*}=z_k^{\ell^*}+\alpha\,v^{\ell^*,p}_{u,r},
  \label{eq:steering}
\end{equation}
where $\alpha$ controls the magnitude and sign of the intervention. The addition leaves the model parameters unchanged; $\alpha=0$ recovers the unsteered model. The new prompt need not contain the historical sessions or an explicit user profile.

\section{Experiments}
\label{sec:experiments}

In this section, we conduct experiments on \upshift{} to answer the following questions:
\begin{enumerate}[label=\textbf{Q\arabic*.},leftmargin=4em,labelwidth=1.8em,labelsep=.6em,itemsep=4pt,topsep=4pt]
  \item \textbf{Representation:} Can the proposed pipeline effectively represent user preferences?
  \item \textbf{Personalization:} Does the same user prefer steered responses to unsteered responses?
  \item \textbf{Generalization:} Does the effect transfer to topics and tasks absent from the extraction history?
  \item \textbf{Specificity:} Does user $u$ prefer $v_u$ over a direction extracted from another user?
  \item \textbf{Causality:} Is $+\hat{v}_u$ more effective than $-\hat{v}_u$?
\end{enumerate}

\subsection{Experimental Setup}
We evaluate \upshift{} in two complementary settings. First, we use the PrefEval benchmark \citep{zhaos2025} to test whether steering vectors can represent and control explicitly specified preferences, isolating the representational capacity of the method from the problem of inferring preferences from interaction histories (Q1). Second, we evaluate the complete pipeline using 40 simulated users constructed from public LinkedIn profiles. These profiles contain objective biographical and professional information but no explicit descriptions of subjective response preferences.

For the second setting, we randomly sample daily trending topics collected from Google and use a lightweight language model to reformulate them into natural conversation starters. A profile-conditioned simulated user then interacts with either Qwen3-8B or Qwen3-32B, while the user profile remains hidden from the assistant. We collect the resulting conversations and apply \upshift{} to extract a preference direction for each user.

We evaluate the extracted directions on a separate set of newly sampled topics. For each user--topic pair, we generate an unsteered baseline response and multiple steered responses that vary in extraction configuration, steering coefficient, and intervention sign. The responses are randomly paired under a balanced matching schedule and evaluated through blind A/B comparisons. DeepSeek-V4-Pro acts as the profile-conditioned preference judge, with response order randomized and the generation configuration concealed. We aggregate the pairwise outcomes into Elo ratings for each configuration. This evaluation tests personalization and cross-topic generalization (Q2--Q3), while other-user directions and negative-direction interventions provide controls for specificity and directionality (Q4--Q5).

\subsection{Preference Representation}

To answer Q1, we use PrefEval to test whether the steering vector represents preference information strongly enough to improve preference control. Each of the benchmark's classification and generation modes contains 1,000 examples spanning 20 topics. An example first states a user preference, either explicitly or implicitly, then inserts an extended sequence of unrelated dialogue before posing a target query. The task therefore tests whether the model can recover and follow the original preference despite interference from the intervening context. Because these intervening turns are benchmark distractors rather than a coherent trajectory of repeated feedback from one user, we use the \textsc{First}-- RM configuration instead of extracting the final response after all intervening turns. Following the benchmark taxonomy, we group the 20 topics into seven domains and report micro-averaged accuracy within each domain. Classification accuracy measures whether the model identifies the target preference, whereas generation accuracy is assigned by the benchmark's preference judge.

Table~\ref{tab:prefeval} reports the results for Qwen3-8B. Without steering, the model achieves 78.4\% classification accuracy and 68.8\% judged generation accuracy overall. With \upshift{} at $\alpha=0.02$, the corresponding accuracies are 78.5\% and 71.7\%. Increasing the intervention strength to $\alpha=0.05$ yields the best overall result in both modes: 83.6\% for classification and 76.4\% for generation, absolute gains of 5.2 and 7.6 percentage points over the unsteered baseline. At $\alpha=0.07$, performance remains close to the optimum but decreases slightly to 83.4\% and 76.1\%, respectively. The non-monotonic trend indicates that stronger intervention is not always better and that $\alpha$ should be selected using held-out validation data.

\begin{table}[t]
\caption{\textbf{PrefEval accuracy (\%) by preference domain.} The best result in each column within each evaluation mode is bold.}
\label{tab:prefeval}
\centering
\small
\setlength{\tabcolsep}{3pt}
\begin{tabular*}{\linewidth}{@{\extracolsep{\fill}}llccccccccc@{}}
\toprule
Model & Method & $\alpha$ & Travel & Shop & Life. & Entert. & Pet & Educ. & Prof. & Overall\\
\midrule
\multicolumn{11}{l}{\textit{Classification}}\\
Qwen3-8B & Baseline & -- & 72.4 & 81.1 & 80.1 & 78.7 & 74.4 & 81.2 & 86.1 & 78.4\\
Qwen3-8B & +\upshift{} & .02 & 72.9 & 79.9 & 80.3 & 78.8 & 76.7 & 85.3 & \textbf{87.7} & 78.5\\
Qwen3-8B & +\upshift{} & .05 & \textbf{77.3} & \textbf{86.1} & 85.9 & \textbf{85.4} & 79.4 & 88.1 & 87.3 & \textbf{83.6}\\
Qwen3-8B & +\upshift{} & .07 & \textbf{77.3} & 85.4 & \textbf{86.2} & 84.2 & \textbf{81.5} & \textbf{88.4} & 86.7 & 83.4\\
\midrule
\multicolumn{11}{l}{\textit{Generation (preference judge)}}\\
Qwen3-8B & Baseline & -- & 65.4 & 67.9 & 69.7 & 70.1 & 72.1 & 70.6 & 72.2 & 68.8\\
Qwen3-8B & +\upshift{} & .02 & 64.2 & 70.0 & 73.7 & 74.2 & 81.4 & 71.8 & 77.8 & 71.7\\
Qwen3-8B & +\upshift{} & .05 & 69.4 & \textbf{75.0} & \textbf{77.5} & 80.1 & \textbf{82.7} & \textbf{78.8} & 81.0 & \textbf{76.4}\\
Qwen3-8B & +\upshift{} & .07 & \textbf{70.1} & 73.7 & \textbf{77.5} & \textbf{80.9} & 79.5 & 78.4 & \textbf{81.1} & 76.1\\
\bottomrule
\end{tabular*}
\end{table}

\subsection{Personalization and Generalization}
To evaluate personalization and cross-topic generalization (Q2--Q3), we construct 40 simulated users from profiles of real individuals. For each user and assistant model (Qwen3-8B and Qwen3-32B), a profile-conditioned LLM conducts 60 four-round conversations on randomly sampled topics. We apply each extraction configuration of \upshift{} to these interaction histories, producing a separate preference direction for every user.

The full-scale evaluation uses ten newly sampled topics per user, none of which appear in that user's extraction conversations. For each topic, we generate responses from the unsteered baseline and from the model steered by different configurations of the user's own preference direction. DeepSeek-V4- Pro, conditioned on the corresponding user profile, performs blind pairwise preference judgments with four possible outcomes: prefer A, prefer B, prefer both, or prefer neither. We randomize the A/B order to mitigate position bias and use a balanced matching schedule so that all configurations participate in approximately the same number of comparisons. Each configuration receives approximately 60 judgments per user for each model.

\begin{figure}[t]
  \centering
  \includegraphics[width=\linewidth]{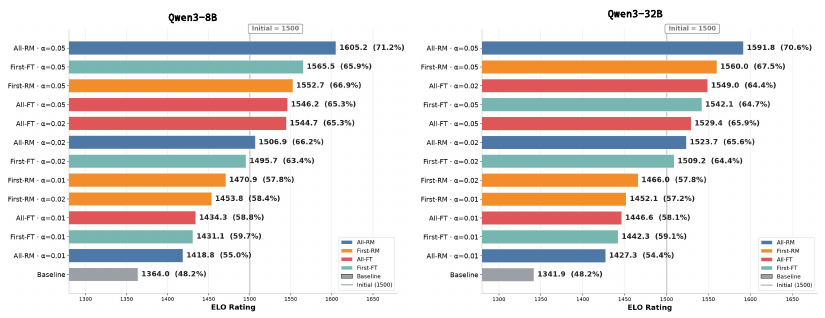}
  \caption{\textbf{Preference alignment on held-out topics over 40 simulated users.} The left and right panels report results for Qwen3-8B and Qwen3-32B, respectively. Bars show the macro-averaged Elo rating across users for each extraction configuration and steering coefficient, ordered by Elo; parenthesized values report the corresponding mean pairwise win rates. The dashed line marks the common initial Elo of 1500; the unsteered baselines obtain Elo ratings of 1364.0 and 1341.9 for Qwen3-8B and Qwen3-32B, respectively. Elo uses $K=32$ and an initial rating of 1500; each configuration participates in approximately 60 comparisons per user for each model.}
  \label{fig:alignment}
\end{figure}

Figure~\ref{fig:alignment} summarizes the results over all 40 simulated users evaluated on held-out topics. We compute the Elo ratings separately for each user and report their macro-average for each extraction configuration and steering coefficient, preventing users with more valid comparisons from disproportionately affecting the aggregate result. For both Qwen3-8B and Qwen3-32B, every tested steering configuration achieves a higher mean rating than the corresponding unsteered baseline. \textsc{All}--RM with $\alpha=0.05$ performs best for both models: its mean Elo rating is 1605.2 on Qwen3-8B, compared with 1364.0 for the baseline, and 1591.8 on Qwen3-32B, compared with 1341.9 for the baseline. These results provide evidence that preference directions estimated from interaction histories can improve alignment with the same user on previously unseen topics. They also show that performance is sensitive to the combination of extraction configuration and steering strength.

\subsection{User Specificity and Directionality}

We next test whether the performance improvement is attributable to the extracted preference direction itself and its sign, rather than to a generic activation intervention. For user specificity (Q4), under the same extraction configuration and steering coefficient, we compare the Elo rating obtained using user $u$'s matched preference direction with the mean rating obtained using preference directions sampled from mismatched users. We randomly select five target users and, for each target user, sample directions from five other users as mismatched controls; we then average the resulting user-level contrasts. For directionality (Q5), we compare positive and negative injection of the same user's preference direction. We evaluate all 40 users separately, with each A/B preference test comparing only the positively and negatively steered responses and excluding the unsteered baseline, and then average the user-level results. Let $E(u,v)$ denote the Elo rating assigned by the judge conditioned on user $u$, and let $\mathcal{U}_u^-$ denote the set of mismatched control users. For each user, we define the two paired contrasts
\begin{equation}
  \Delta_{\mathrm{spec}}(u)=E(u,v_u)-\frac{1}{|\mathcal{U}_u^-|}\sum_{u'\in\mathcal{U}_u^-}E(u,v_{u'}),\quad
  \Delta_{\mathrm{dir}}(u)=E(u,+v_u)-E(u,-v_u),
  \label{eq:contrasts}
\end{equation}

We compute each contrast at the user level and report its macro-average over the five target users for Q4 and over all 40 users for Q5. Positive values of $\Delta_{\mathrm{spec}}$ indicate that the judge prefers the correctly matched user direction, while positive values of $\Delta_{\mathrm{dir}}$ indicate that the learned orientation is more effective than its negation.

\begin{figure}[t]
  \centering
  \includegraphics[width=\linewidth]{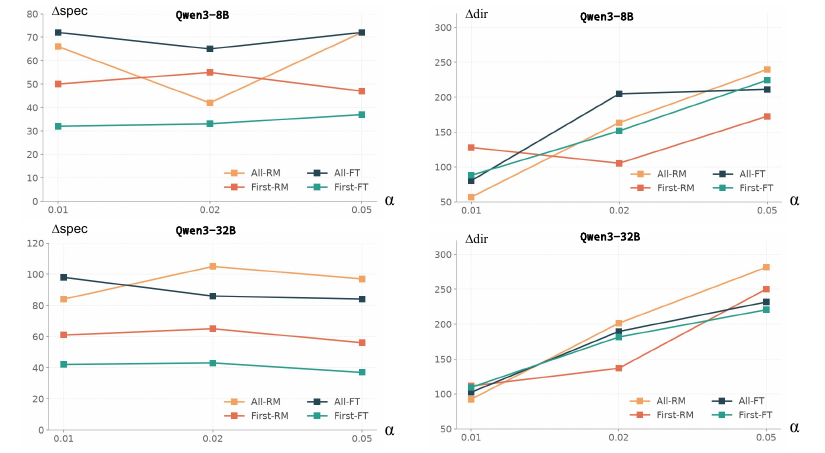}
  \caption{\textbf{User specificity and directionality controls.} The top and bottom rows report Qwen3-8B and Qwen3-32B, respectively. The left column shows the macro-averaged specificity contrast $\Delta_{\mathrm{spec}}$, comparing each target user's matched direction with directions from mismatched users; the right column shows the macro-averaged directionality contrast $\Delta_{\mathrm{dir}}$, comparing positive and negative injection of the same user's direction. Values are Elo-rating differences, and positive values favor the matched user direction or the learned orientation. Lines connect results from the same extraction configuration across steering coefficients $\alpha$.}
  \label{fig:controls}
\end{figure}

Figure~\ref{fig:controls} reports both controls. Every tested configuration produces a positive specificity contrast for both model sizes and all three steering coefficients: the matched-user direction exceeds the mismatched-user controls by approximately 32--72 Elo points on Qwen3-8B and 37--105 points on Qwen3-32B. The magnitude does not increase monotonically with $\alpha$, suggesting that user identity is encoded consistently across the tested intervention strengths but that stronger steering does not necessarily improve discrimination between users.

The directionality contrasts are also positive throughout, showing that the learned sign is essential rather than arbitrary. Unlike specificity, directionality generally strengthens as $\alpha$ increases. At $\alpha=0.05$, \textsc{All}--RM yields the largest positive-versus-negative gap for both models, approximately 240 Elo points on Qwen3-8B and 282 points on Qwen3-32B. This agrees with the held-out personalization results in Figure~\ref{fig:alignment}, where the same configuration and coefficient perform best. Together, these controls indicate that the gains arise from a user-matched, correctly oriented preference direction rather than from injecting an arbitrary activation vector.

\section{Limitations and Ethical Considerations}
\label{sec:limitations}

\paragraph{Limits of the implicit-preference hypothesis.} Our hypothesis is statistical rather than a guarantee about every turn. Continued interaction need not imply improvement, and later feedback may introduce a new requirement rather than refine a persistent preference. Preferences may also be task-dependent or inconsistent across sessions. Aggregation isolates a preference only when some component recurs; otherwise, it may cancel conflicting signals or encode task-specific progression.

\paragraph{Compactness versus capacity.} A single vector is storage- and prompt-efficient but has limited capacity for multiple, conditional, or nonlinear preferences and cannot select among conflicting preferences based on the task. Multiple directions or context-dependent routing may address this limitation.

\paragraph{Preference may conflict with safety objectives.} Preference alignment need not improve truthfulness, calibration, or safety: users may favor unwarranted confidence, sycophancy, or unsafe content. Such preferences should not be optimized unconditionally, even when they improve measured win rates.

\paragraph{Access and portability.} \upshift{} requires white-box activation access and forward-pass intervention, precluding closed text-only APIs. Directions are specific to the checkpoint, architecture, layer, and representation space, and may require re-estimation after model updates rather than transfer across architectures.

\section{Conclusion}
\label{sec:conclusion}

We introduced \upshift{}, a training-free method that converts hidden-state changes in human--AI interaction histories into reusable user-specific steering directions. By aggregating the displacement between initial and feedback-conditioned responses across sessions, \upshift{} personalizes new responses without placing historical conversations in context or updating model parameters. Experiments on PrefEval show that these directions can represent and control preference information, while evaluations with 40 simulated users show improved preference alignment on held-out topics for both Qwen3-8B and Qwen3-32B. Matched-user and positive-versus-negative controls further confirm that the improvements depend on the user-specific direction and its learned orientation. Overall, the results support the implicit-preference hypothesis and demonstrate that ordinary dialogue traces can provide compact supervision for lightweight personalization without explicit preference profiles.

\bibliography{references}
\bibliographystyle{iclr2026_conference}
\end{document}